\documentclass{article}

\PassOptionsToPackage{numbers, compress}{natbib}

\usepackage[preprint]{main}

\usepackage[utf8]{inputenc} 
\usepackage[T1]{fontenc}    
\usepackage{hyperref}       
\usepackage{url}            
\usepackage{booktabs}       
\usepackage{amsfonts}       
\usepackage{nicefrac}       
\usepackage{microtype}      
\usepackage{xcolor}         
\usepackage{amsmath} 
\usepackage{graphicx}
\usepackage{tabularx}
\usepackage{multirow}
\usepackage{float}
\usepackage{siunitx}
\title{Estimating prevalence with precision and accuracy}

\author{%
  Aime Bienfait Igiraneza,
  Christophe Fraser,
  Robert Hinch \\
  Pandemic Sciences Institute \\
  University of Oxford \\
  Oxford, UK \\
  \texttt{\small\{aime.igiraneza, christophe.fraser, robert.hinch\}@ndm.ox.ac.uk} \\
}

\begin{document}

\maketitle

\begin{abstract}

    Unlike classification, whose goal is to estimate the class of each data point, quantification (or prevalence estimation) aims to estimate the distribution of classes in a dataset. An important task in prevalence estimation is to quantify the uncertainty in prevalence estimates. In this paper, we introduce \textit{Precise Quantifier} (PQ), a Bayesian aggregative quantifier that achieves narrow prediction intervals with sufficient coverage (\textit{i.e.,} sufficient proportion of intervals containing the true prevalence). We find that PQ produces more precise prevalence estimates than existing methods as the discriminative power of the underlying classifier increases and as the validation-to-test size ratio increases. These empirical results suggest that PQ uses validation information more effectively to quantify uncertainty in prevalence estimates than existing approaches.\footnote{https://github.com/BDI-pathogens/rPQ}
    
\end{abstract}

\section{Introduction}

Consider an ongoing epidemic caused by a fast-evolving virus, where there exists an effective monoclonal antibody treatment.
After a few months, the number of treatment failures starts to increase due to selection for resistant viral strains.
A natural question arises: what is the current prevalence of viral resistance to the monoclonal antibody?
Using laboratory assays to test for antibody resistance in many samples is infeasible due to resource limitations, but sequencing many virus samples is possible.
Therefore, we could use existing samples to train a model that predicts antibody resistance based on viral sequences.
Applying this model to new viral sequences and aggregating the results will provide an estimate of the prevalence of viral resistance to the antibody. However, simply \textit{classifying and counting} will significantly underestimate or overestimate the true resistance prevalence in the new datasets \cite{forman_counting_2005,forman_quantifying_2008}. \textit{Quantification learning} provides a set of methods that adjust for these biases \cite{esuli_learning_2023}.

In quantification learning, the setup has two types of datasets: a \textit{validation} dataset in which every data point is labeled (\textit{e.g.,} each viral sequence has a neutralization phenotype associated with it) and \textit{test} datasets in which data points have no labels. The task consists in estimating the prevalence of the label of interest in test datasets, given the validation dataset. Here we focus on cases where labels are classes, although quantification extends to learning problems with other types of labels \cite{bella_aggregative_2014}. The quantification task differs from the classification one in terms of both objectives and assumptions \cite{gonzalez_why_2017}. In classification, models aim to classify individual inputs, assuming that test sets are as independently and identically distributed as the datasets used to train the classifier. However, in quantification, models aim to estimate class prevalence, assuming that this prevalence varies between test sets and datasets used to train the quantifier. Although classifying is not a prerequisite to quantifying, here we focus on aggregative quantification methods, which produce prevalence estimates based on individual predictions from a classifier \cite{esuli_learning_2023}. 

In addition to a point estimate of the prevalence, it is important to quantify the uncertainty of the estimate. A common method is bootstrapping point estimates from existing quantifiers, yielding bootstrap confidence intervals (BCIs) \cite{tasche_confidence_2019}. Alternatively, Bayesian analysis has been used to naturally get uncertainty quantification in posterior distributions of the prevalence \cite{amos_when_2008,ziegler_bayesian_2024, keith_uncertainty-aware_2018, denham_gain--lose-_2021, fiksel_generalized_2022}. It is not clear which approach (Bayesian or non-Bayesian) achieves the highest precision ({\it i.e.}, the narrowest intervals) with sufficient coverage ({\it i.e.}, with repeat experiments, the proportion of intervals containing the true prevalence reaches the confidence level). Through a series of simulations, \citet{tasche_confidence_2019} suggested that BCIs could have enough precision with sufficient coverage. However, \citet{ziegler_bayesian_2024} recently proposed a Bayesian method, referred to as BayesianCC hereafter, which seemed to yield credible intervals that are tighter than BCIs, without compromising coverage. 

In this paper, we introduce \textit{Precise Quantifier} (PQ), a new Bayesian aggregative quantifier, which achieves high precision with sufficient coverage. We compare PQ to existing quantifiers on a range of simulated and real-world datasets. We find that PQ produces more precise prevalence estimates than existing methods as the discriminative power of the underlying classifier increases and as the validation-to-test size ratio increases.

\section{Precise Quantifier -- PQ}
\label{theory}
In this section, we formalize the quantification problem, while specifically setting out the assumptions used to derive PQ in the context of Bayesian statistics. 
We focus on the binary classification problem, where the true class is either 0 (negative class) or 1 (positive class), denoted by the binary variable $Y$. Throughout, we use three types of datasets: the training dataset to train the underlying classifier; the validation dataset to train the quantifier; and the test dataset for which we estimate prevalence. 
Data points in the test set are not labeled with classes, while training and validation data points are labeled.
We define $f(X)$ as the probabilistic classifier trained using the training dataset, where $X$ is the descriptive data used for classification.
The value of $f(X)$ is the predicted probability that a sample is in the positive class.

The input to PQ is the validation set of labeled data $\mathcal{V} = \{(f(x^V_1),y^V_1), ..., (f(x^V_{n_V}),y^V_{n_V})\}$ of size $n_V$, and the test set of unlabeled data $\mathcal{T} = \{f(x^T_1), ..., f(x^T_{n_T})\}$ of size $n_T$.
Additionally, PQ has one hyperparameter $N_{\rm bin }$, which is the number of bins used to group the data. We will say more about $N_{\rm bin }$ later.  
The validation set is partitioned into a subset containing only positive samples ($\mathcal{V_+}$) and one containing only negative samples ($\mathcal{V_-}$).
Similarly, the test set can be partitioned into $\mathcal{T_+}$ and $\mathcal{T_-}$ respectively; however, since we are not given the class labels in the test set, we cannot do this \textit{a priori}.
Finally, we define $P_V$ as the probability of an event in the validation set and $P_T$ as the probability of an event in the test set.

PQ's key assumption is that the classifier's estimates for positive samples in the validation and test sets are drawn from the same underlying distribution, and the same for negative samples. 
Mathematically this is, $P_V(f(X)|Y) = P_T(f(X)|Y)$, which is called the \textit{weak prior probability shift assumption} \cite{lipton_detecting_2018}.
This is a relaxation of the prior probability shift assumption stating that $P_V(X|Y) = P_T(X|Y)$ \cite{amos_when_2008}. 
However, in general the prevalence is different in the validation and test datasets ({\it i.e.}, $P_V(Y) \neq P_T(Y)$), which implies $P_V(f(X)) \neq P_T(f(X))$.
The problem of estimating prevalence in the test dataset can then be formulated in a multi-level Bayesian framework, where the underlying distributions of the positive and negative classes ({\it i.e.}, $P(f(X)|Y=1)$ and $P(f(X)|Y=0)$) are learned simultaneously. Therefore, the posterior distribution of the prevalence includes the uncertainty from both validation and test data sets, allowing principled uncertainty quantification. 

At its core, PQ estimates $P(f(X)|Y=1)$ and $P(f(X)|Y=0)$, the underlying distributions of the classifier's predictions for positive and negative samples. 
Note that the classifier's predictions will be biased by the prevalence in the training dataset, with this effect being especially strong for samples which the classifier struggles to classify.
In particular, for samples where the descriptive data $X$ provides no additional information on the class, the probabilistic classifier $f(X)$ will simply be the prevalence in the training dataset.
However, typically there will be some samples that the classifier will assign to a class with high confidence, regardless of the prevalence in the training dataset.
Therefore, the distribution of the classifier's estimates will often not be well captured by a standard probability distribution ({\it e.g.}, a beta distribution), which is why we use non-parametric distributions for the classifier's estimates. Specifically, we bin the classifier’s positive class probabilities into $N_{\rm bin}$ bins spanning [0, 1]. This binning is reminiscent of the approach used in distribution matching methods such as HDy and DyS \cite{gonzalez-castro_estimating_2010,maletzke_dys_2019}. However, existing methods use fixed binning (\textit{i.e.}, equal-width intervals). In our analyses, we advocate for quantile binning (\textit{i.e.}, binning such that bins have equal amount of validation data points) because we found that this strategy is more robust to systematic biases in the classification scores than fixed binning (see section \ref{quantile_binning}). Throughout this paper, our notation specifies whether quantile binning was used by adding a "q" suffix, while the number of bins is also indicated as a suffix. For example, when $N_{\rm bin}=8$, we write PQ\_8 and PQ\_8\_q for fixed and quantile binning, respectively.

The first part of the model is to assume that each element in $\mathcal{V_+}$ ($\mathcal{V_{-}}$) is independently drawn from a distribution containing the whole positive (negative) class, so that
\begin{equation}
    \mathcal{\bar{V}_+} \sim {\rm Multinomial}(n_V^+,\{p^+_1,..,p^+_{N_{\rm bin }}\})\quad {\rm and }\quad
    \mathcal{\bar{V}_-} \sim {\rm Multinomial}(n_V^-,\{p^-_1,..,p^-_{N_{\rm bin }}\}),
\end{equation}
where $\mathcal{\bar{V}_\pm} = \{ v^\pm_1,.\, ., v^\pm_{N_{bin}}\}$ are the number of samples in each bin, $n_V^\pm$ are the sizes of $\mathcal{V_\pm}$, and $\{p^\pm_1,..,p^\pm_{N_{\rm bin }}\}$ are parameters which we estimate.
Similarly, using the \textit{weak prior probability shift assumption},
the elements in $\mathcal{T_\pm}$ are drawn from the same distributions as  $\mathcal{V_\pm}$, so that
\begin{equation}
    \mathcal{\bar{T}_+} \sim {\rm Multinomial}(n_T^+,\{p^+_1,..,p^+_{N_{\rm bin }}\})\quad {\rm and }\quad
    \mathcal{\bar{T}_-} \sim {\rm Multinomial}(n_T^-,\{p^-_1,..,p^-_{N_{\rm bin }}\}),
\end{equation}
where  $\mathcal{\bar{T}_\pm}=\{t^{\pm}_1,..,t^{\pm}_{N_{\rm bin }}\}$ are the number of samples in each bin, and $n_T^\pm$ are the sizes of $\mathcal{T_\pm}$.
However, the samples in $\mathcal{T}$ are unlabeled, so we do not know the split between  $\mathcal{T_\pm}$, just the total number of samples in each bin, which we define as $\mathcal{\bar{T}}=\{t_1,..,t_{N_{\rm bin }}\}$ and $n_T$ as the total number of samples in $\mathcal{T}$.

We now estimate the prevalence in a multi-level Bayesian framework. 
Recalling that $f(x^T_i) \in \mathcal{T}$, we let $f(x^T_i)$, the classifier's estimate, be in the $k^{\rm th}$ bin. We also recall that $y^T_i$ is the corresponding (unknown) true class for $x^T_i$ (that is, if $x^T_i$ is in the negative class then $y^T_i=0$, and $y^T_i=1$ if $x^T_i$ is in the positive class). 
Using the \textit{weak prior probability shift assumption}, the likelihood of $f(x^T_i)$ conditioned on $y^T_i$ is
\begin{equation}
P(f(x^T_i)|y^T_i,p^+_k,p^-_k) = y^T_i p^+_k + ( 1 - y^T_i)p^-_k.
\end{equation}

We now introduce a prior distribution on $y^T_i$ 
\begin{equation}
y^T_i \sim {\rm Bernoulli}(\theta_{\rm pr}),
\end{equation}
where $\theta_{\rm pr}$ is the prior prevalence, so that
\begin{align*}
P(f(x^T_i)| \theta_{\rm pr},p^+_k,p^-_k ) 
& = P(f(x^T_i)|y^T_i =0,p^+_k,p^-_k ) P(y^T_i=0|\theta_{\rm pr})\\
& + P(f(x^T_i)|y^T_i =1,p^+_k,p^-_k ) P(y^T_i=1|\theta_{\rm pr}) \\
& =\theta_{\rm pr} p^+_k + ( 1 - \theta_{\rm pr})p^-_k,    
\end{align*}
and 
\begin{equation}
P(y^T_i=1| \theta_{\rm pr},p^+_k,p^-_k ) = \frac{\theta_{\rm pr}  p_k^+}{  \theta_{\rm pr} p_k^+ + (1-\theta_{\rm pr})p_k^- }.
\end{equation}
The prior prevalence $\theta_{\rm pr}$ will be one of the quantities we estimate from the data. 
Note, the prior prevalence is an estimated parameter in the posterior distribution ({\it i.e.}, this is a multi-level Bayesian method).
Assuming that every element in $\mathcal{T}$ is an independent sample, we get the log-likelihood
\begin{equation}
\log \left\{ P(\mathcal{T}|\theta_{\rm pr},\{p^+\} ,\{p^-\} )\right\} = \sum_{k=1}^{N_{\rm bin}} t_k \log \bigl( \theta_{\rm pr} p^+_k + ( 1 - \theta_{\rm pr})p^-_k \bigr),
\end{equation}
where we recall that $t_k$ is the number of samples in $\mathcal{T}$ in the $k^{\rm th}$ bin. 
Assuming that samples in the validation and test set are independent, we get
\begin{multline}
\log \left\{ P(\mathcal{T},\mathcal{V}|\theta_{\rm pr},\{p^+\} ,\{p^-\} ) \right\} \\= \sum_{k=1}^{N_{\rm bin}} \left\{ t_k \log \bigl( \theta_{\rm pr} p^+_k + ( 1 - \theta_{\rm pr})p^-_k \bigr) + v^+_k \log p^+_k + v^-_k \log p^-_k\right\} + c,
\end{multline}
where $c$ is a constant which is independent of the parameters.
Using Bayes' theorem, we have the posterior distribution 
\begin{equation}
\label{eqn_posterior}
P(\theta_{\rm pr},\{p^+\} ,\{p^-\} | \mathcal{T},\mathcal{V})  \propto P(\mathcal{T},\mathcal{V}|\theta_{\rm pr},\{p^+\} ,\{p^-\} ) P(\theta_{\rm pr},\{p^+\} ,\{p^-\} ),
\end{equation}
where $P(\theta_{\rm pr},\{p^+\} ,\{p^-\} )$ is the prior distribution on the parameters.
By default we use independent uninformative priors for each parameter ({\it i.e.}, $\theta_{\rm pr} \sim {\rm Uniform}(0,1)$ and $\{p^\pm\} \sim {\rm Dirichlet}(1)$) .
Finally, the test sample prevalence $\theta$ is given by
\begin{equation}
\theta = \frac{1}{n_{T}} \sum_{i=1}^{n_{T}} y^T_i.
\end{equation}
Thus, assuming independence of each sample in $\mathcal{T}$
\begin{equation}
\label{eqn_theta}
\theta | \theta_{\rm pr},\{p^+\} ,\{p^-\} \sim \frac{1}{n_{T}}\sum_{k=1}^{N_{\rm bin}} Y_k| \theta_{\rm pr},\{p^+\} ,\{p^-\} ,
\end{equation}
where
\begin{align}
Y_k | \theta_{\rm pr},\{p^+\} ,\{p^-\} 
& \sim {\rm Binomial} \left( t_k,P(y^T_i=1| \theta_{\rm pr},p^+_k,p^-_k ) \right) \\
& \sim {\rm Binomial} \left( t_k,\frac{\theta_{\rm pr}  p_k^+}{  \theta_{\rm pr} p_k^+ + (1-\theta_{\rm pr})p_k^- } \right).
\end{align}
To obtain the full posterior distribution for $\theta|\mathcal{T},\mathcal{V}$, we integrate over the posterior distribution of $\left\{\theta_{\rm pr},\{p^+\} ,\{p^-\}\right\}|\mathcal{T},\mathcal{V}$. The posterior distribution for $\left\{\theta_{\rm pr},\{p^+\} ,\{p^-\}\right\}|\mathcal{T},\mathcal{V}$ (Equation ~\ref{eqn_posterior}) is sampled using the default MCMC sampler in Stan. 
Note that this involves sampling a small number ($2N_{\rm Bin} + 1$) of continuous parameters, which is very efficient compared to directly sampling $\{y^T_1,..,y^T_{n_T}\}$,
which is a large number of discrete parameters and thus difficult to sample.
To sample from the posterior distribution of $\theta|\mathcal{T},\mathcal{V}$, for each posterior sample of $\left\{\theta_{\rm pr},\{p^+\} ,\{p^-\}\right\}|\mathcal{T},\mathcal{V}$, we sample a single value of $\theta$ using Equation \ref{eqn_theta}. 
Our experimental results were calculated using four Markov chain Monte Carlo (MCMC) chains, each with 1000 posterior samples and 1000 warm-up samples.

Our audience will realize that our approach makes a distinction between $\theta_{\rm pr}$ and $\theta$. $\theta_{\rm pr}$ estimates the prevalence in the underlying population from which the test set was sampled, while $\theta$ estimates the realized prevalence in the test sample. We argue that $\theta$ is the estimate aligned with the quantification evaluation protocol that uses test samples of finite size. Thus, we evaluate PQ using $\theta$, and we use the term "prediction interval" (PI) to mean the central interval of the posterior predictive distribution containing the desired probability mass (\textit{e.g.}, 90\%).

\section{Experimental setup}
\label{experiments}
\subsection{Simulated datasets}
\label{simulated datasets}
To evaluate quantifiers in controlled settings, we simulated quantification problems with varying difficulty levels. Specifically, we used two normal distributions with unit variance \cite{tasche_confidence_2019}, with one mean always set to 0, while the other mean was set to 0.5, 1, 1.5, 2, or 2.5. In each scenario, the normal distribution centered at 0 represented the negative class, while the other normal distribution represented the positive class. Note that the theoretical area under the receiver operating characteristic curve (AUC) for these simulations is given by $\Phi(\frac{\mu}{\sqrt{2}})$, where $\mu$ is the mean for the positive-class distribution and $\Phi$ is the standard normal cumulative distribution function. Thus, the five simulated scenarios correspond to problems with theoretical AUCs of 0.64, 0.76, 0.86, 0.92, and 0.96, respectively, with values rounded to two decimal places. For each of the scenarios, we drew a training sample of 1000 data points with positive class prevalence of 50\%. We then fitted a logistic regression model on this set. The validation data used to train quantifiers was a separate sample of varying size depending on the experiment, with positive class prevalence of 50\%. We emphasize that this validation set was different from the logistic model's training set. To test quantifiers, we drew test samples with varying class prevalence, following the protocol outlined in Section \ref{protocol}.

\subsection{Antibody datasets}
\label{antibody_datsets}
We ran quantifiers on two anti-HIV-1 broadly neutralizing antibodies: 10-1074 and 3BNC117 \cite{mouquet_complex-type_2012,scheid_sequence_2011}. All data was downloaded from CATNAP (April 2026) \cite{yoon_catnap_2015}. The raw data consisted of aligned amino acid sequences of Env with corresponding IC80 values (\textit{i.e.,} concentrations of the antibody needed to reduce {\it in vitro} infectivity by 80\%). Since we aimed to perform binary quantification, we binarized IC80 values using a \SI{1}{\micro\gram/\milli\liter} threshold. For cases with multiple IC values for the same Env-antibody pair, the geometric mean value was used. Right-censored values automatically resulted in resistance cases. In summary, an Env sequence was resistant to an antibody if IC80 $\geq$ \SI{1}{\micro\gram/\milli\liter} or if any of the IC values was right-censored. Left-censored IC values were transformed to the corresponding numerical part ({\it i.e.}, "<$x$" became "$x$"). We one-hot encoded sequences before fitting models. That is, each character was represented as a vector of all zeros except a 1 at the index of that character in the vocabulary. We included the gap character "-" in the vocabulary, since deletions and insertions may be informative \cite{zacharopoulou_calculated_2022}. For any other character that was neither an amino acid nor "-", all elements in the encoding vector were zeros. 

After preprocessing the data, we fitted logistic regression models with L2 regularization and hyperparameters \texttt{C} and \texttt{max\_iter} set to 0.1 and 2000, respectively. All other hyperparameters were left to their default values in scikit-learn v1.6.1 \cite{pedregosa_scikit-learn_2011}. Classification performance is reported in Table \ref{tab:table 3.3}. For quantification, we used class probabilities predicted via five-fold cross-validation evaluation to avoid overconfident predictions. That is, each prediction was made using a model whose training dataset did not include the data point in question. The size of the dataset used to train quantifiers ({\it i.e.}, the validation set) was set to 250, with the same positive class prevalence as in the corresponding dataset ({\it i.e.}, $p$ in Table \ref{tab:table 3.3}). The size of the test sample was also set to 250.

\begin{table}[H]
\centering
\caption[Antibody datasets and classification performance]{\textbf{Antibody datasets and classification performance.} MCC was calculated using thresholds that yielded maximum values ({\it i.e.}, maximum MCC is shown). $n$ is size. $p$ is positive class prevalence. Numbers between parentheses are standard deviations. Decimal values are rounded to two decimal places.}
\begin{tabular}{lcccc}
\toprule
\textbf{Antibody} & \textbf{AUC} & \textbf{MCC} & \textbf{$n$} & \textbf{$p$}\\
\midrule
10-1074 & 0.85 (0.03) & 0.58 (0.04) & 798 & 0.48\\
3BNC117 & 0.86 (0.01) & 0.58 (0.02) & 937 & 0.46\\
\bottomrule
\end{tabular}
\label{tab:table 3.3}
\end{table}

\subsection{Review datasets}
We also tested quantification methods on three review datasets from QuaPy \cite{moreo_quapy_2021}. Specifically, these are the hp, kindle, and imdb datasets.\footnote{https://hlt-isti.github.io/QuaPy/manuals/datasets.html} The datasets' statistics are summarized in Table \ref{tab:table 3.4}. Using training datasets as given in QuaPy, we trained logistic regression models, with \texttt{max\_iter}, \texttt{class\_weight}, and \texttt{solver} set to $1000$, "balanced", and "liblinear", respectively in scikit-learn v1.6.1 \cite{pedregosa_scikit-learn_2011}. All the other hyperparameters were left to their default values. We report classification performance in Table \ref{tab:table 3.5}. For each of the three cases, datasets used to train and test quantifiers (the validation and test sets, respectively) were sampled from the test data given in QuaPy ($n_{test}$ in Table \ref{tab:table 3.4}). The validation sample size was set to 250 with 50\% positive class prevalence, while the test sample size was set to either 250 or 1000, depending on the experiment.

\begin{table}[!ht]
 \caption[Review datasets as published in QuaPy]{\textbf{Review datasets as published in QuaPy.} $n_{train}$ is the size of QuaPy training set. $p_{train}$ is the prevalence of the positive class in QuaPy training set. $n_{test}$ is the size of QuaPy test set. $p_{test}$ is the prevalence of the positive class in QuaPy test set.}
  \centering
  \begin{tabular}{lllll}
    \toprule
    Dataset     & $n_{train}$     & $p_{train}$    & $n_{test}$    &  $p_{test}$\\
    \midrule
    hp & 9533  & 98.2\% & 18399       & 93.5\%\\
    kindle     & 3821 & 91.9\% & 21591     & 93.7\%\\
    imdb     & 25000  & 50\% & 25000      & 50\%\\
    \bottomrule
  \end{tabular}
  \label{tab:table 3.4}
\end{table}

\begin{table}[!ht]
 \caption[Classification performance on review datasets]{\textbf{Classification performance on review datasets.} These metrics were calculated on QuaPy test dataset. MCC was calculated using thresholds that yielded maximum values ({\it i.e.}, maximum MCC is shown). All numbers are rounded to two decimal places.}
 
  \centering
  \begin{tabular}{lllll}
    \toprule
    Dataset     & AUC & MCC\\
    \midrule
    hp & 0.91 & 0.44\\
    kindle & 0.95 & 0.58 \\
    imdb & 0.95 & 0.77\\
    \bottomrule
  \end{tabular}
  \label{tab:table 3.5}
\end{table}

\subsection{LeQua datasets}
\label{LeQua_datasets}
Finally, we considered three datasets from LeQua competitions: T1A from LeQua 2022 \cite{esuli_detailed_2022}, and T1 and T4 from LeQua 2024 \cite{esuli_overview_2024}. These datasets were also accessed via QuaPy \cite{moreo_quapy_2021}. For each of the three datasets, we trained logistic regression models on half of the training data ($n=2500$), with \texttt{max\_iter} set to 1000 and all other hyperparameters set to their default values in scikit-learn v1.6.1 \cite{pedregosa_scikit-learn_2011}. We then used the other half as validation data ($n=2500$) to train quantifiers. The classification performance on this validation data is reported in Table \ref{tab:table 4}. Note that the split between training and validation sets was stratified such that both sets have the same class prevalence as the original dataset. This positive class prevalence in the validation set was 78\%, 77\%, and 81\% for T1, T1A, and T4, respectively. Quantifiers were tested using what is called "validation samples" and "test samples" in QuaPy. For each of the three datasets, there is a total of 6000 such samples, each of size 250 and with class prevalence that covers the [0, 1] range.

\begin{table}[!ht]
 \caption[Classification performance on LeQua datasets]{\textbf{Classification performance on LeQua datasets.} These metrics were calculated on half of the training dataset, with models being trained on the other half. MCC was calculated using thresholds that yielded maximum values ({\it i.e.}, maximum MCC is shown). All numbers are rounded to two decimal places.}
 
  \centering
  \begin{tabular}{lllll}
    \toprule
    Dataset     & AUC & MCC\\
    \midrule
    T1A & 0.88 & 0.57\\
    T1 & 0.91 & 0.63 \\
    T4 & 0.93 & 0.67\\
    \bottomrule
  \end{tabular}
  \label{tab:table 4}
\end{table}

\subsection{Evaluation protocol for quantifiers}
\label{protocol}
Unlike classification, quantification requires that methods be evaluated with prevalence shifts between validation and test sets. In our experiments, we adopted the artificial-prevalence protocol (APP) \cite{forman_counting_2005}, with preset test sample sizes. For the simulated datasets, sampling test sets of a given size and class prevalence was straightforward as explained in Section \ref{simulated datasets}. For the antibody and review datasets, we preset test sizes such that we could sample without replacement. Except for LeQua datasets, for which test sets were already given by QuaPy as described in Section \ref{LeQua_datasets}, reported results were based on 1010 prevalence estimates, corresponding to 1010 test sets generated in the following way:
\begin{itemize}
    \item Discretize the interval [0,1] into 101 evenly-spaced values, which we consider to be the positive class prevalences.
    \item For each class prevalence, sample 10 test sets, each generated without replacement.
\end{itemize}

Overall, we compared PQ to BayesianCC, EMQ, HDy, DyS, PACC, T50, X, ACC, MAX, MS, and KDEyML \cite{ziegler_bayesian_2024,bella_quantification_2010,gonzalez-castro_estimating_2010,saerens_adjusting_2002,forman_counting_2005,forman_quantifying_2008,forman_quantifying_2006,maletzke_dys_2019,moreo_kernel_2024}. We provide further details on these methods in the \ref{quantifiers}. These are aggregative quantifiers that cover a wide range of approaches and are easily accessible in the QuaPy framework that we used \cite{moreo_quapy_2021}. We refer to \citet{esuli_learning_2023} for a comprehensive review of existing quantification methods, including those that are non-aggregative. 

For PQ and BayesianCC, we use the terms "prediction interval" (PI) and "credible interval" (CrI), respectively, to mean the central interval of their posterior prevalence distributions containing the desired probability mass (90\% in our case). For all the other quantifiers, which output point estimates, bootstrap confidence intervals (BCIs) were generated as follows:
\begin{itemize}
    \item For each of 1000 iterations, draw with replacement a sample from the validation set, without changing its size and class proportion. In addition, draw with replacement a sample from the test set, maintaining its size and class proportion. 
    \item Train the quantifier on the validation sample and estimate class prevalence on the test sample.
    \item Calculate 90\% percentile confidence intervals with the 1000 class prevalence estimates. 
\end{itemize}

Throughout our analyses, we report three metrics, namely the "width" of 90\% PIs or CrIs or BCIs, their "coverage" ({\it i.e.}, the proportion of the intervals containing the true prevalence), and the "absolute error", which is the absolute difference between predicted prevalence and true prevalence. For PQ and BayesianCC, we use the mean of the posterior distribution as the predicted prevalence. For all three metrics, we also report 95\% confidence intervals calculated using 1000 bootstrap samples (these are not to be confused with the 90\% BCIs, PIs or CrIs).

\section{Results}

\subsection{Quantile binning is robust to calibration bias in classification scores}
\label{quantile_binning}
Binning classification scores is an important aspect of PQ, along with methods such as HDy and DyS \cite{gonzalez-castro_estimating_2010,maletzke_dys_2019}. Traditionally, this binning is done using fixed intervals (\textit{e.g.}, 4 bins with edges 0, 0.25, 0.5, 0.75, and 1). However, we expected this strategy to be sensitive to calibration bias, while quantile binning (\textit{i.e.}, binning such that there is an equal amount of data in each bin) would be robust to this bias. 

\begin{figure}[!ht]
    \centering
    \includegraphics[width=1\linewidth]{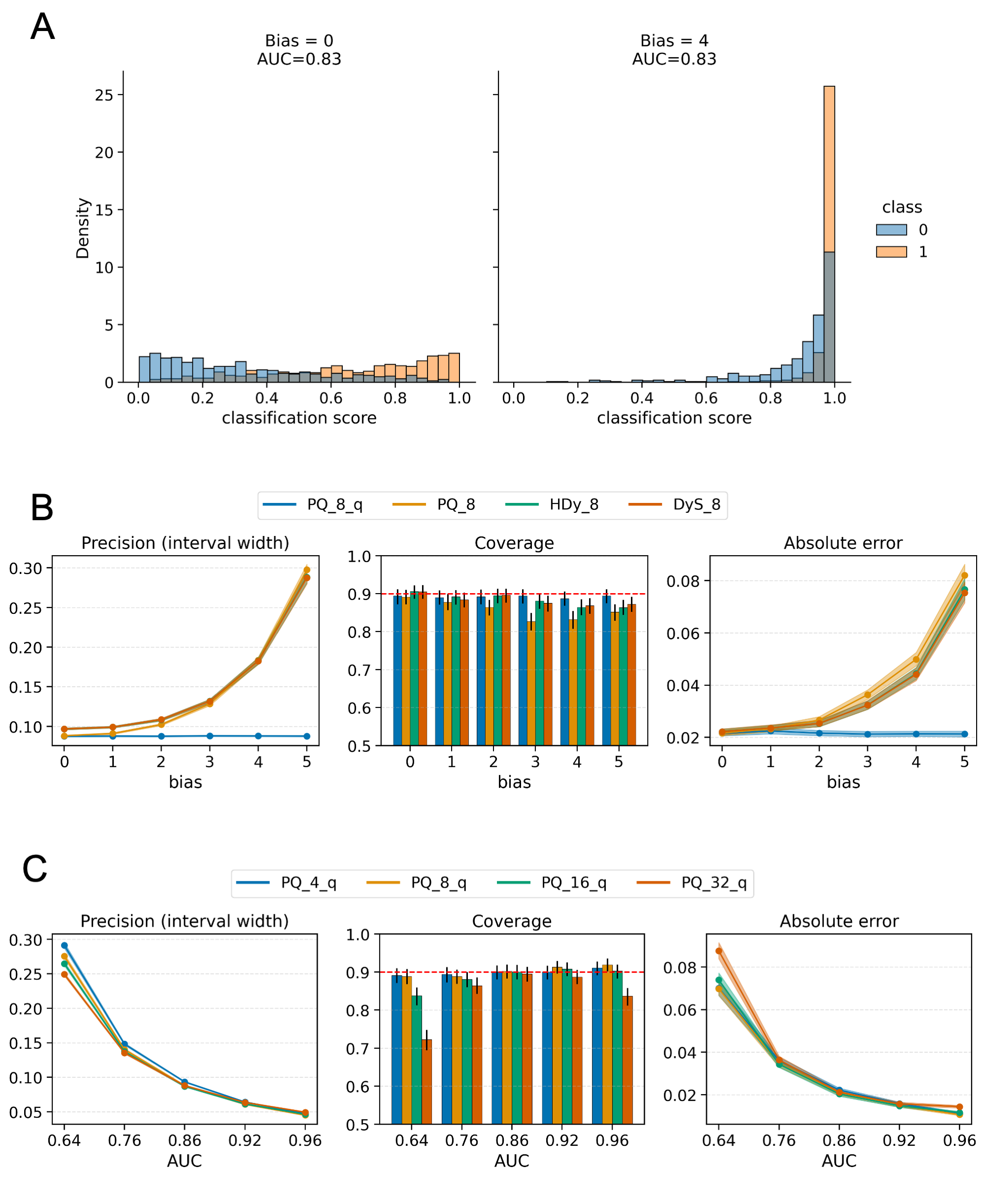}
    \caption{\textbf{Binning}. \small (\textbf{A}) The simulated calibration bias moves predicted probabilities towards the positive class, without changing the ranking of predictions (\textit{i.e.}, AUC stays the same). In subfigures \textbf{B} and \textbf{C}, precision and coverage are for 90\% prediction intervals for PQ and 90\% bootstrap confidence intervals for the other methods. The dashed red lines in the coverage plots indicate the nominal coverage level of 90\%. Numbers in names of quantifiers indicate the number of used bins, while the "q" suffix indicates that quantile binning was used. (\textbf{B}) As calibration bias increases, fixed binning increases absolute error and decreases precision and coverage. By calibration bias, we mean $b$ in $\sigma(\operatorname{logit}(\hat p_i)+b)$, where $\sigma$ is the logistic function and $\hat p_i=f(x_i)$ is the original predicted positive class probability for input $x_i$. (\textbf{C}) Using too many bins risks compromising coverage without any significant gains in terms of precision and absolute error. In all cases, the size of the validation and test sets was set to 1000, with 50\% positive class prevalence in the validation set. The means of the negative and positive class distributions were 0 and 1.5, respectively (\textit{i.e.}, theoretical $AUC=0.86$). Mean values are shown; error bars and error bands represent 95\% bootstrap confidence intervals, calculated with 1000 samples.}
    \label{fig:Fig_1}
\end{figure}

To study robustness to systematic calibration bias, we transformed the classifier's predicted positive-class probabilities in logit space. Specifically, for an original predicted probability $\hat p_i=f(x_i)$, we defined the biased score
\[
\hat p_i^{(b)}=\sigma(\operatorname{logit}(\hat p_i)+b),
\]
where $\sigma$ is the logistic function and $b$ controls the strength of the bias. We hereafter may simply refer to $b$ as "bias". This transformation is monotone, and therefore does not change the ranking of samples or the AUC, but it systematically shifts predicted probabilities toward the positive class when $b>0$ (Fig. \ref{fig:Fig_1}A).

As calibration bias increased (\textit{i.e.}, as $b$ increased from 0 to 5), there was an increase in the width of prediction intervals for PQ with fixed binning and the width of bootstrap confidence intervals for HDy and DyS (PQ\_8, HDy\_8, and DyS\_8 in Fig. \ref{fig:Fig_1}B). At the same time, when fixed binning was used, absolute errors of the three methods increased as bias increased (Fig. \ref{fig:Fig_1}B). Moreover, the increase in absolute error corresponded to a decrease in coverage, as the calibration bias increased. In contrast, in terms of precision, coverage and absolute error, PQ with quantile binning (PQ\_8\_q in Fig. \ref{fig:Fig_1}B) was not affected by the increase in bias.

An alternative to binning is to assume parametric distributions over the classification scores, the approach taken by KDEyML for instance \cite{moreo_kernel_2024}. Given the reliance of such approaches on the geometry of the classification scores, we expected them to fail as calibration bias increased. Indeed, we found that KDEyML's performance in terms of coverage and absolute error degraded as bias increased (Fig. \ref{fig:Fig_S1}). Furthermore, under extreme calibration bias, precision and absolute error for BayesianCC and ACC worsened substantially, but without compromising coverage (Fig. \ref{fig:Fig_S1}). All the other tested quantifiers maintained their performance despite changes in bias (Fig. \ref{fig:Fig_S1}). 

We concluded that quantile binning is more robust to calibration bias in classification scores than fixed binning. Furthermore, our results confirmed that the assumption of distributions in the classification scores may compromise quantification performance as calibration bias increases.

One remaining challenge is determining the number of bins ($N_{\rm bin }$ in our description of PQ in Section \ref{theory}). Across simulation scenarios, increasing $N_{\rm bin }$ from 4 to 8 improved PQ's precision and absolute error without compromising coverage (Figs. \ref{fig:Fig_1}C and \ref{fig:Fig_S2}). However, further increases to 16 and 32 could compromise coverage, especially in very weak and very powerful classification settings (Fig. \ref{fig:Fig_1}C). We note that when $N_{\rm bin }$ was set to 16 or 32 with medium classification power (\textit{i.e.}, AUC = 0.86), coverage was not compromised even with relatively small validation and test sets ($n=250$; Fig. \ref{fig:Fig_S2}). However, such high $N_{\rm bin }$ did not provide significant improvement in terms of precision and absolute error (Fig. \ref{fig:Fig_S2}). Thus, we concluded that small values for $N_{\rm bin }$ were enough and by default we set $N_{\rm bin }$ to 8.

\subsection{PQ achieves high precision without compromising accuracy}

\begin{figure}[!t]
    \centering
    \includegraphics[width=1\linewidth]{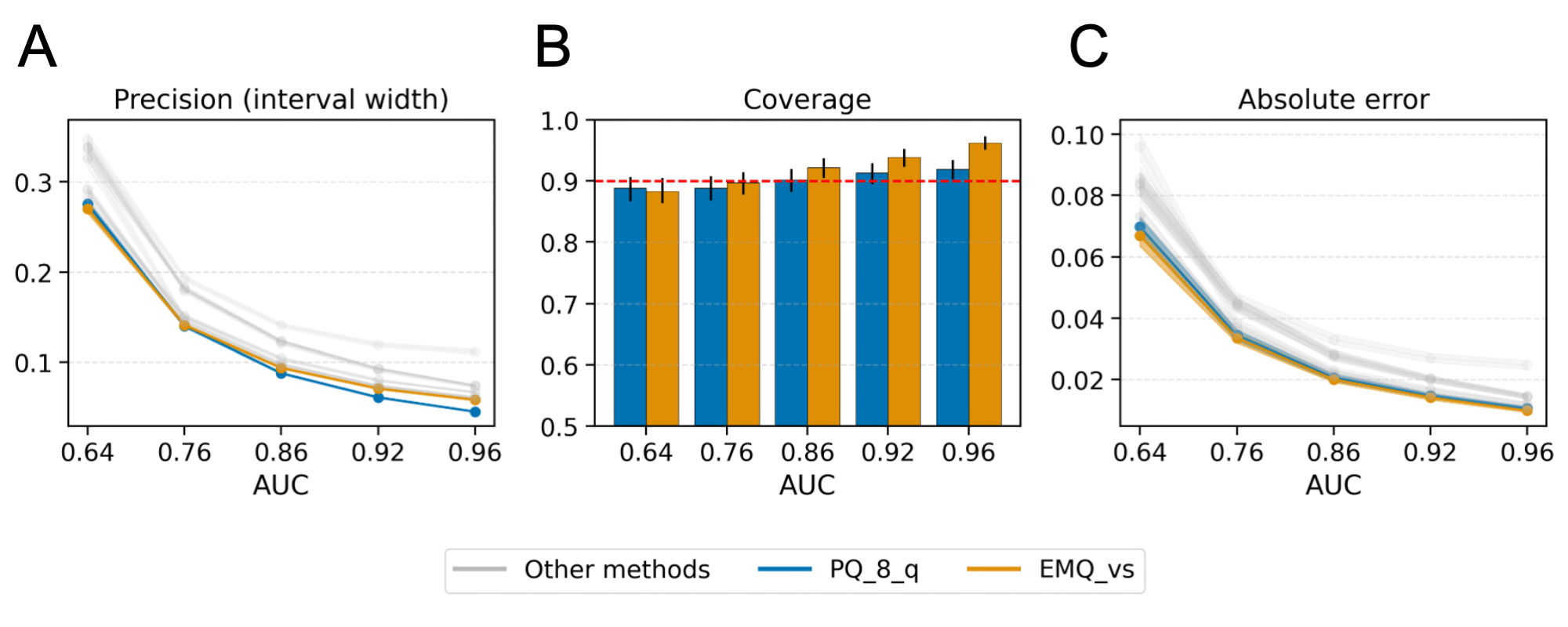}
    \caption{\textbf{Impact of discriminative power}. \small As classification power increases, PQ becomes more precise than existing approaches. Precision and coverage are for 90\% prediction intervals for PQ and 90\% bootstrap confidence intervals for EMQ. PQ\_8\_q is PQ with 8 quantile bins. EMQ\_vs is EMQ with vector scaling used as the calibration method. The dashed red line in subfigure \textbf{B} indicates the nominal coverage level of 90\%. Validation and test sizes were fixed at 1000, with 50\% positive class prevalence in the validation set. Mean values are shown; error bars and error bands represent 95\% bootstrap confidence intervals, calculated with 1000 samples.}
    \label{fig:Fig_2}
\end{figure}

As classification power increased, precision and absolute error improved across quantifiers (Figs. \ref{fig:Fig_1}C and \ref{fig:Fig_2}). However, we found that, without compromising coverage, PQ became more precise than all the other quantifiers as the underlying classifier became very powerful (\textit{i.e.}, $AUC\geq0.86$; Fig. \ref{fig:Fig_2}A). EMQ was the second best method in terms of precision, while both EMQ and PQ had similar absolute errors regardless of the discriminative power of the classifier.

\begin{figure}[!ht]
    \centering
    \includegraphics[width=1\linewidth]{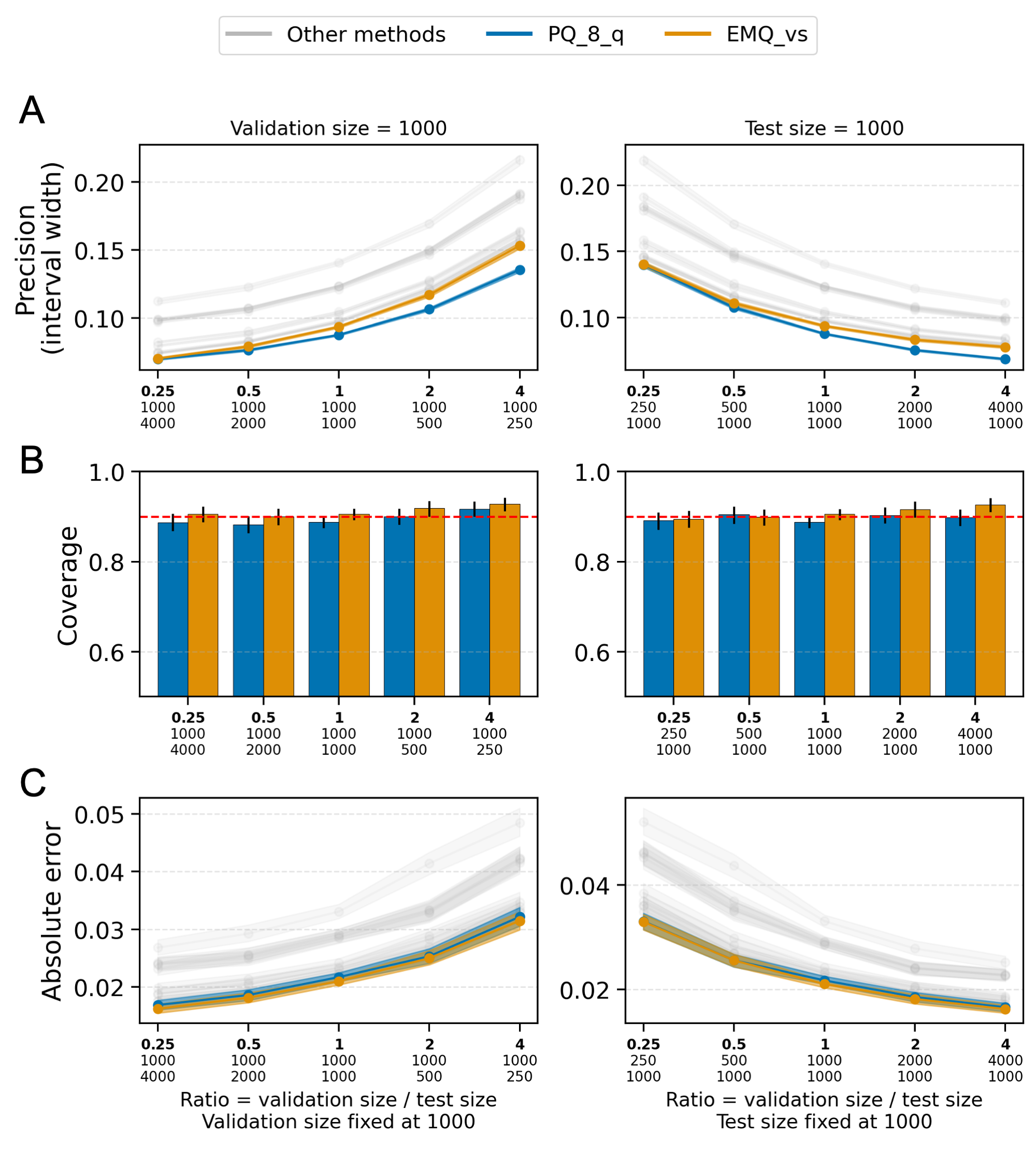}
    \caption{\textbf{Impact of validation-to-test size ratio}. \small As the validation-to-test size ratio increases, PQ becomes more precise than existing approaches. Precision and coverage are for 90\% prediction intervals for PQ, 90\% credible intervals for BayesianCC, and 90\% bootstrap confidence intervals for the other methods. The dashed red lines in subfigure \textbf{B} indicate the nominal coverage level of 90\%. The tick labels on the x-axis are shown in three rows: the first row is for the ratio, the second is for the validation size, and the third is for the test size. Left column represents the case when only test size is changing from 4000 to 250, with validation size fixed at 1000. The right column is for the case when only the validation size is changing from 250 to 4000, with test size fixed at 1000. In all cases, the positive class prevalence in the validation set was fixed at 50\%. In all cases, the means of the negative and positive class distributions were 0 and 1.5, respectively (\textit{i.e.}, theoretical $AUC=0.86$). PQ\_8\_q is PQ with 8 quantile bins. EMQ\_vs is EMQ with vector scaling used as the calibration method. Mean values are shown; error bars and error bands represent 95\% bootstrap confidence intervals, calculated with 1000 samples.}
    \label{fig:Fig_3}
\end{figure}

As the size of either the validation set or the test set increased, precision and absolute errors improved across quantifiers (Fig. \ref{fig:Fig_3}). However, we also found that, without compromising coverage, PQ became more precise than all other quantifiers as the validation-to-test size ratio increased (\textit{i.e.}, ratio$>0.5$; Fig. \ref{fig:Fig_3}A). EMQ remained the second best method in terms of precision, while both EMQ and PQ maintained similar absolute errors across validation-to-test size ratios.

\begin{figure}[!ht]
    \centering
    \includegraphics[width=1\linewidth]{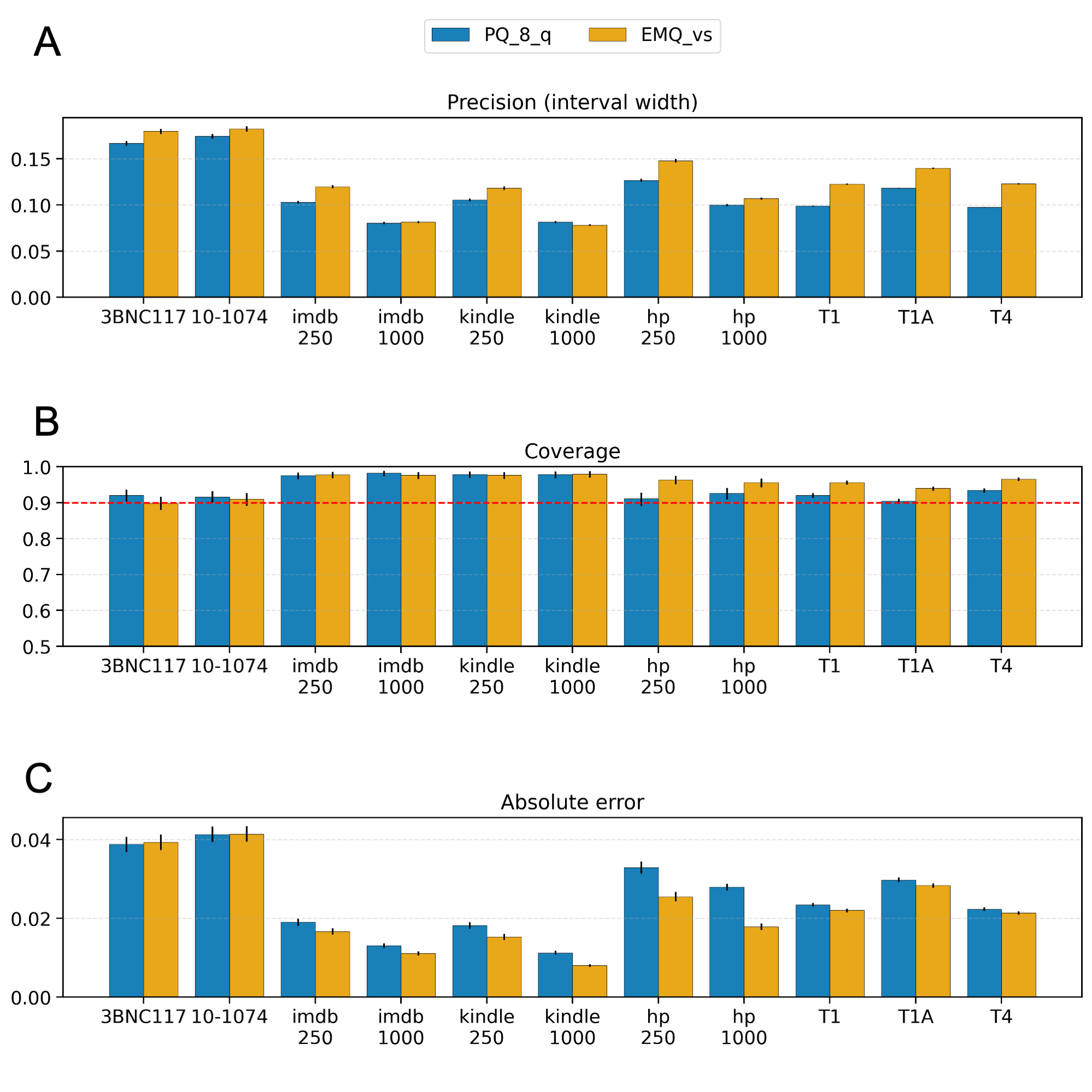}
    \caption{\textbf{Performance on real world datasets}. \small Precision and coverage are for 90\% prediction intervals for PQ and 90\% bootstrap confidence intervals for EMQ. The dashed red line in subfigure \textbf{B} indicates the nominal coverage level of 90\%. The numbers below imdb, kindle and hp on the x-axis indicate the size of the test set. PQ\_8\_q is PQ with 8 quantile bins. EMQ\_vs is EMQ with vector scaling used as the calibration method. Mean values are shown; error bars represent 95\% bootstrap confidence intervals, calculated with 1000 samples. Further details on the datasets and used methods are in Sections \ref{experiments} and \ref{appendix}.}
    \label{fig:Fig_4}
\end{figure}

Across antibody, review, and LeQua datasets, PQ consistently achieved the highest precision without compromising coverage, except for imdb and kindle datasets when validation size and test size were 250 and 1000, respectively (Figs. \ref{fig:Fig_4}, \ref{fig:Fig_S3}--\ref{fig:Fig_S5}). In those two cases, EMQ was at least as precise as PQ, if not slightly more precise than PQ (imdb 1000 and kindle 1000 in Fig. \ref{fig:Fig_4}A). These results are consistent with the simulation results showing the tendency for PQ's precision edge to disappear when the validation-to-test size ratio is low (Fig. \ref{fig:Fig_3}A). In terms of absolute error, EMQ outperformed PQ, except for the antibody datasets, where EMQ and PQ were comparable (Fig. \ref{fig:Fig_4}C). Nonetheless, EMQ was sometimes outperformed by the other methods in terms of absolute error (e.g., KDEyML clearly had the lowest absolute error on hp datasets; Fig. \ref{fig:Fig_S4}). EMQ requires calibrating classification probabilities (see \ref{quantifiers}), and when this calibration is not perfect, EMQ is expected to have higher absolute errors than other methods \cite{alexandari_maximum_2020}.

In summary, these experiments show that PQ becomes more precise than existing approaches as the validation signal improves due to increased discriminative power of the underlying classifier or due to increased validation data relative to test data. We could replicate PQ's precision edge on real world datasets as well, with EMQ maintaining the second best precision, while performing slightly better in terms of absolute error than PQ in several cases. 

\clearpage
\section{Discussion}
\label{sec:Discussion}

We have presented Precise Quantifier (PQ), an aggregative quantification method that achieves high precision with high accuracy (measured in terms of prediction intervals' coverage and absolute difference between estimated prevalence and true prevalence). We have also shown that, in addition to PQ, \citet{saerens_adjusting_2002}'s maximum likelihood method (termed Expectation Maximization Quantifier\textemdash EMQ) coupled with bootstrapping, achieves the second best precision without compromising accuracy. EMQ's high precision is consistent with findings from \citet{tasche_confidence_2019}. 

Several studies have already shown that powerful classifiers allow for more precise (with enough coverage) and less biased prevalence estimates than weak classifiers do \cite{tasche_confidence_2019, ziegler_bayesian_2024, keith_uncertainty-aware_2018, barranquero_quantification-oriented_2015, tasche_minimising_2021}. We arrived at a similar conclusion. Hence, there is value in having a strong classifier, which contrasts with \citet{schumacher_comparative_2025}'s recent suggestion that increasing the discriminative power of the underlying classifier does not substantially reduce absolute error (note that they did not consider precision and coverage as well). Similarly to \citet{schumacher_comparative_2025}'s results, we found that increasing the validation sample size and test size improves quantification precision and accuracy. We showed that PQ's precision edge over the other methods widens as the validation-to-test size ratio increases and as the discriminative power of the underlying classifier increases. These empirical results suggest that PQ uses validation information more effectively than the other approaches to quantify uncertainty in prevalence estimates.

We did not consider analytical ({\it i.e.}, non-bootstrap) confidence intervals \cite{keith_uncertainty-aware_2018, vaz_quantification_2019}, as practitioners are likely to opt for Monte Carlo techniques as they are rather agnostic to specifics of the quantifiers used. A future study should also extend PQ to multi-class cases, since the simple one-versus-all solution ({\it i.e.}, the class of interest being labeled as positive while all the other classes are labeled as negative) has been shown to violate the prior probability shift assumption \cite{donyavi_mc-sq_2024}, despite the original recommendation in Forman's seminal work \cite{forman_quantifying_2008}. Regardless of these limitations, we hope that our analyses clarify key aspects of quantification learning that warrant attention in practice, especially when precision is of interest, in addition to accuracy.

\begin{ack}
We thank Chris Wymant from the Pathogen Dynamics Group in the Pandemic Sciences Institute at the University of Oxford for his helpful feedback and comments on the manuscript. The authors acknowledge financial support from the PANGEA consortium (https://www.pangea-hiv.org/) which was funded by the Bill \& Melinda Gates Foundation (grant INV-007573). The computational aspects of this research were funded from the NIHR Oxford BRC with additional support from the Wellcome Trust Core Award Grant Number 203141/Z/16/Z. The views expressed are those of the author(s) and not necessarily those of the NHS, the NIHR or the Department of Health.
\end{ack}

\bibliographystyle{unsrtnat}  
\bibliography{main}


\setcounter{figure}{0}
\renewcommand{\thefigure}{S\arabic{figure}}
\renewcommand{\theHfigure}{S\arabic{figure}}

\newpage
\begin{figure}[!ht]
    \centering
    \includegraphics[width=1\linewidth]{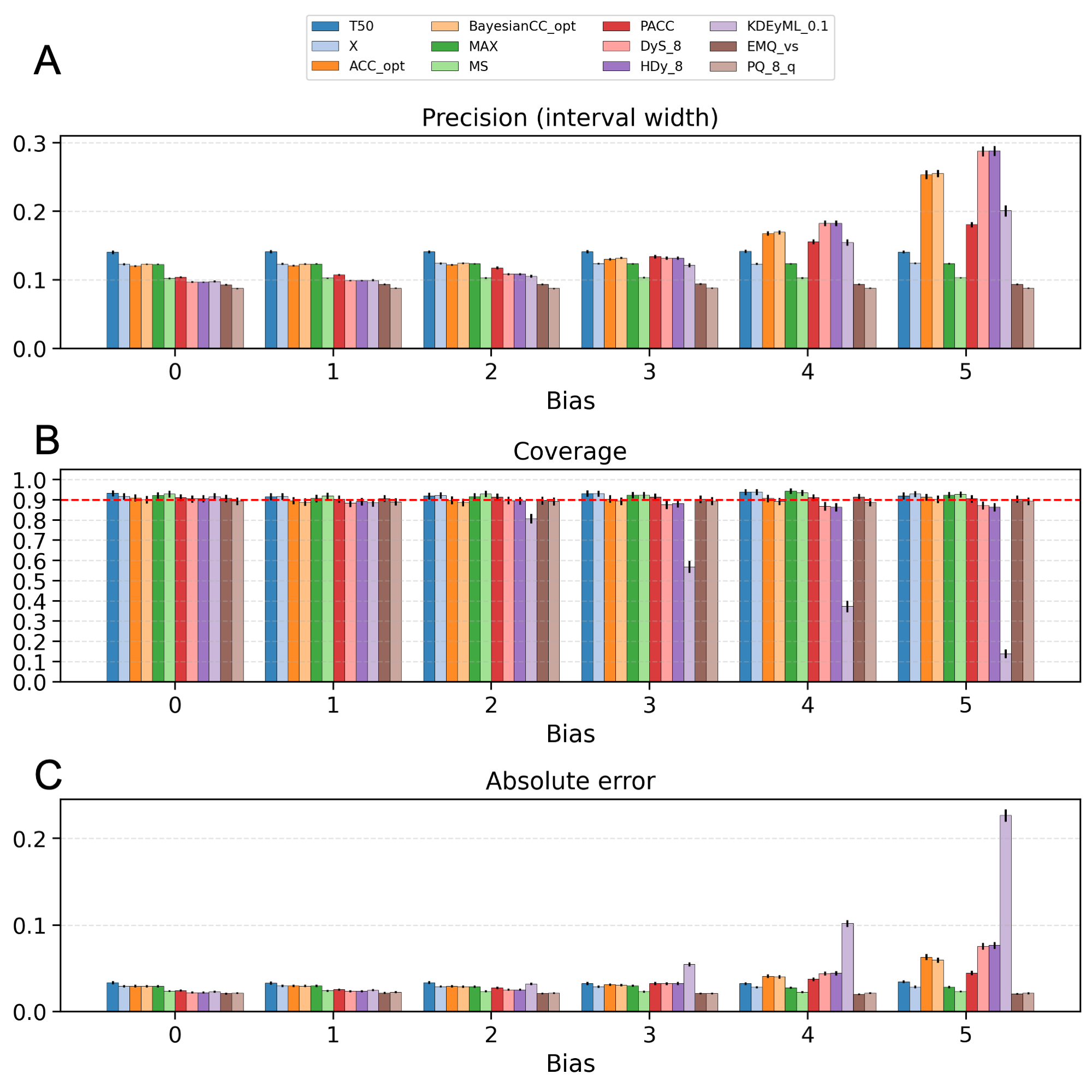}
    \caption{\textbf{Impact of calibration bias on quantification}. \small Precision and coverage are for 90\% prediction intervals for PQ, 90\% credible intervals for BayesianCC, and 90\% bootstrap confidence intervals for the other methods. The dashed red line in subfigure \textbf{B} indicates the nominal coverage level of 90\%. As calibration bias increases, several methods underperform in terms of precision, coverage, and absolute error. For instance, we point out HDy\_8, DyS\_8, ACC\_opt, BayesianCC\_opt, and KDEyML\_0.1. By calibration bias, we mean $b$ in $\sigma(\operatorname{logit}(\hat p_i)+b)$, where $\sigma$ is the logistic function and $\hat p_i=f(x_i)$ is the original predicted positive class probability for input $x_i$. In all cases, the size of the validation and test sets is set to 1000, with 50\% positive class prevalence in the validation set. The means of the negative and positive class distributions are 0 and 1.5, respectively (\textit{i.e.}, theoretical $AUC=0.86$). Mean values are shown; error bars represent 95\% bootstrap confidence intervals, calculated with 1000 samples.}
    \label{fig:Fig_S1}
\end{figure}

\newpage
\begin{figure}[!ht]
    \centering
    \includegraphics[width=1\linewidth]{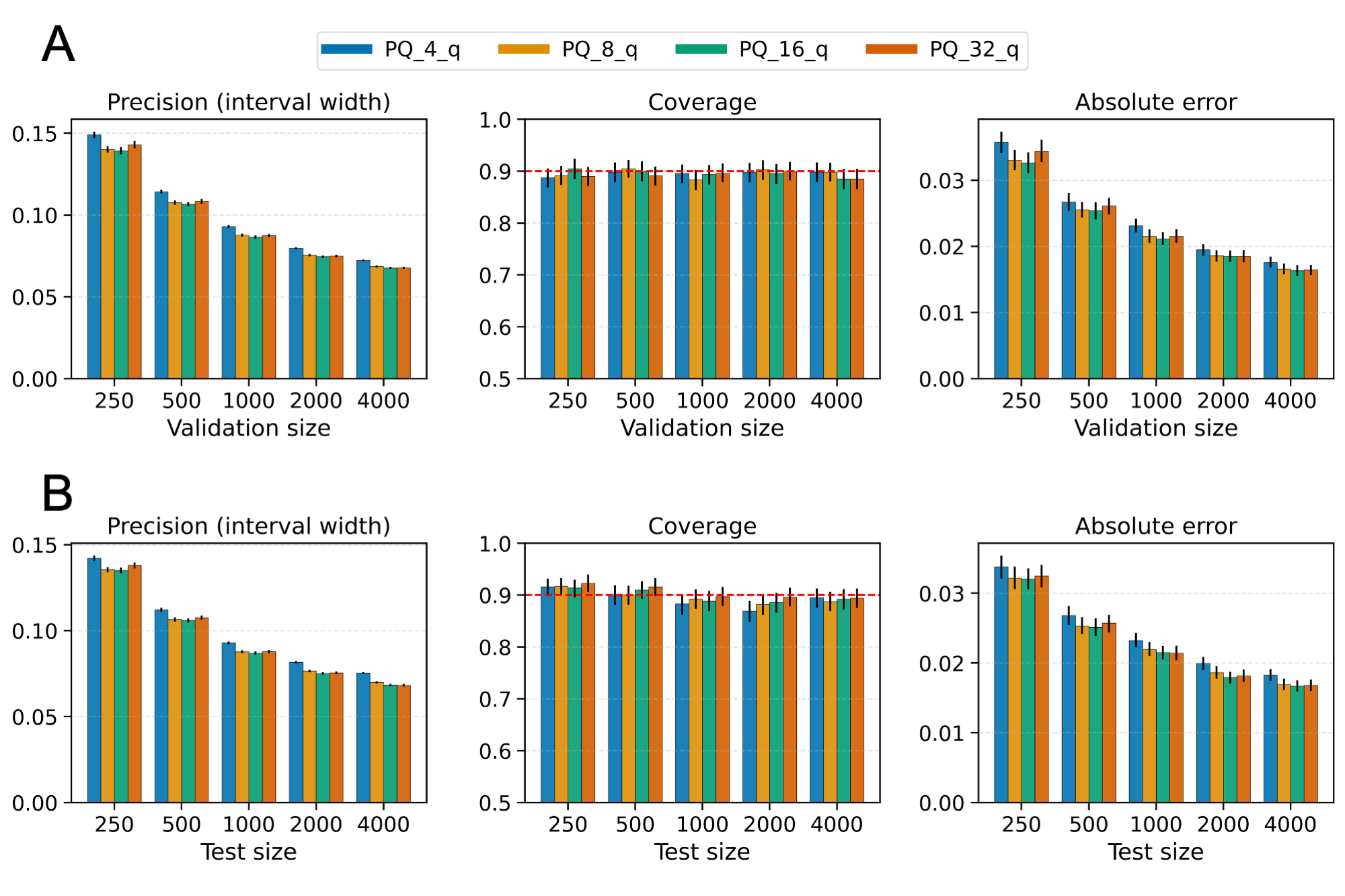}
    \caption{\textbf{Impact of number of bins on PQ}. \small Precision and coverage are for 90\% prediction intervals. The dashed red lines in the coverage plots indicate the nominal coverage level of 90\%. The number in the method name indicates the number of bins, while "q" indicates that quantile binning was used (instead of fixed binning). Increasing the number of bins from 4 to 8 results in slightly lower absolute error and higher precision without comprising coverage. Increase the number of bins beyond 8 does not provide any gains in terms of precision and absolute error. Intriguingly, coverage is not compromised even for the highest number of bins. We note that the means of the negative and positive class distributions were 0 and 1.5, respectively (\textit{i.e.}, theoretical $AUC=0.86$). In Fig. \ref{fig:Fig_1}C, we show that weaker or stronger classification that this one can lead to compromised coverage when 16 and 32 bins are used. (\textbf{A}) Test size was set to 1000. (\textbf{B}) Validation size was set to 1000. In all cases, positive class prevalence in the validation set was set to 50\%. Mean values are shown; error bars represent 95\% bootstrap confidence intervals, calculated with 1000 samples.}
    \label{fig:Fig_S2}
\end{figure}

\newpage
\begin{figure}[!ht]
    \centering
    \includegraphics[width=1\linewidth]{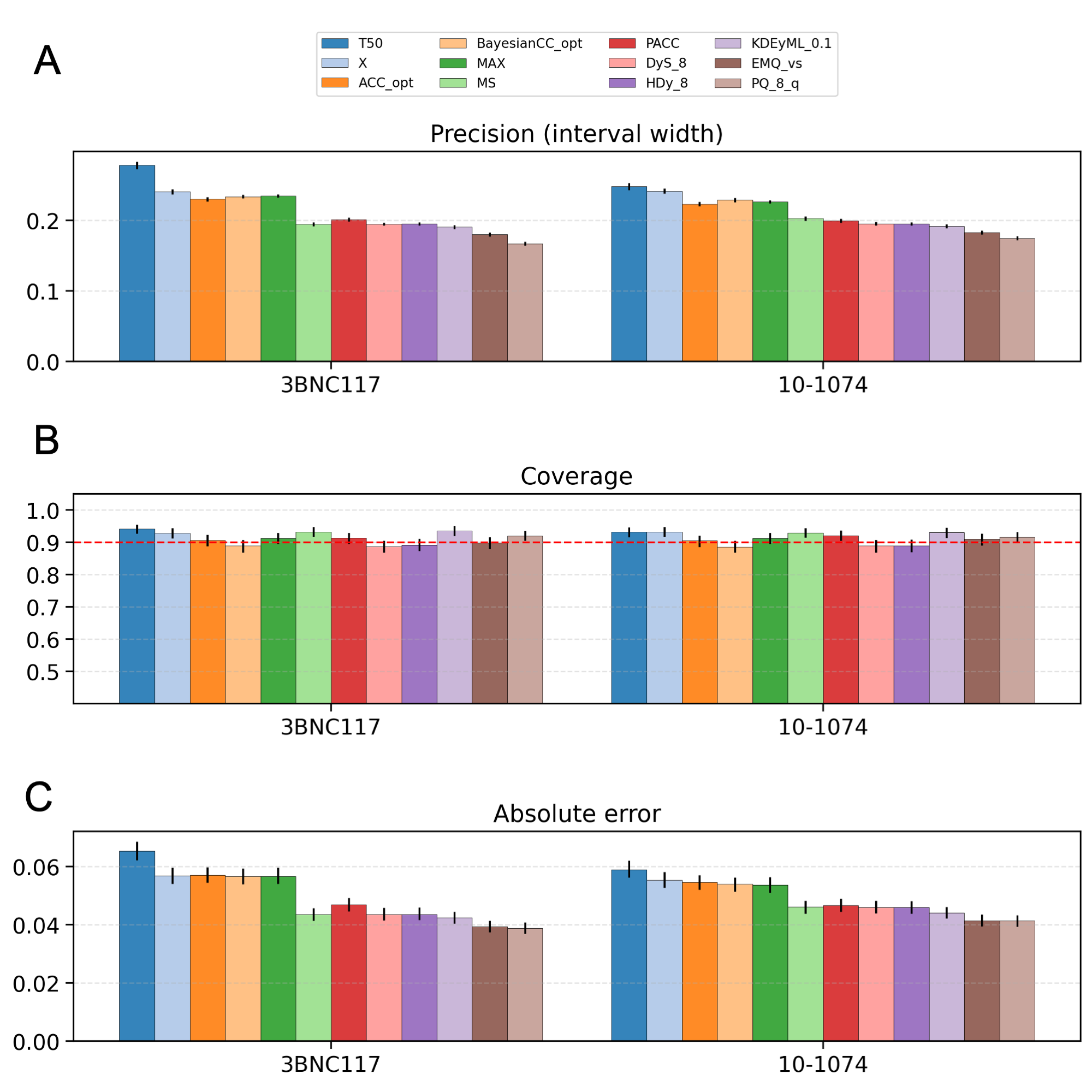}
    \caption{\textbf{Quantification performance on the antibody datasets}. \small Precision and coverage are for 90\% prediction intervals for PQ, 90\% credible intervals for BayesianCC, and 90\% bootstrap confidence intervals for the other methods. The dashed red line in subfigure \textbf{B} indicates the nominal coverage level of 90\%. PQ was more precise than other quantifiers; EMQ was a close second in terms of precision; however, in terms of absolute error, both PQ and EMQ were comparable. Mean values are shown; error bars represent 95\% bootstrap confidence intervals, calculated with 1000 samples. Further details on the datasets and underlying methods are in Sections \ref{experiments} and \ref{appendix}.}
    \label{fig:Fig_S3}
\end{figure}

\newpage
\begin{figure}[!ht]
    \centering
    \includegraphics[width=1\linewidth]{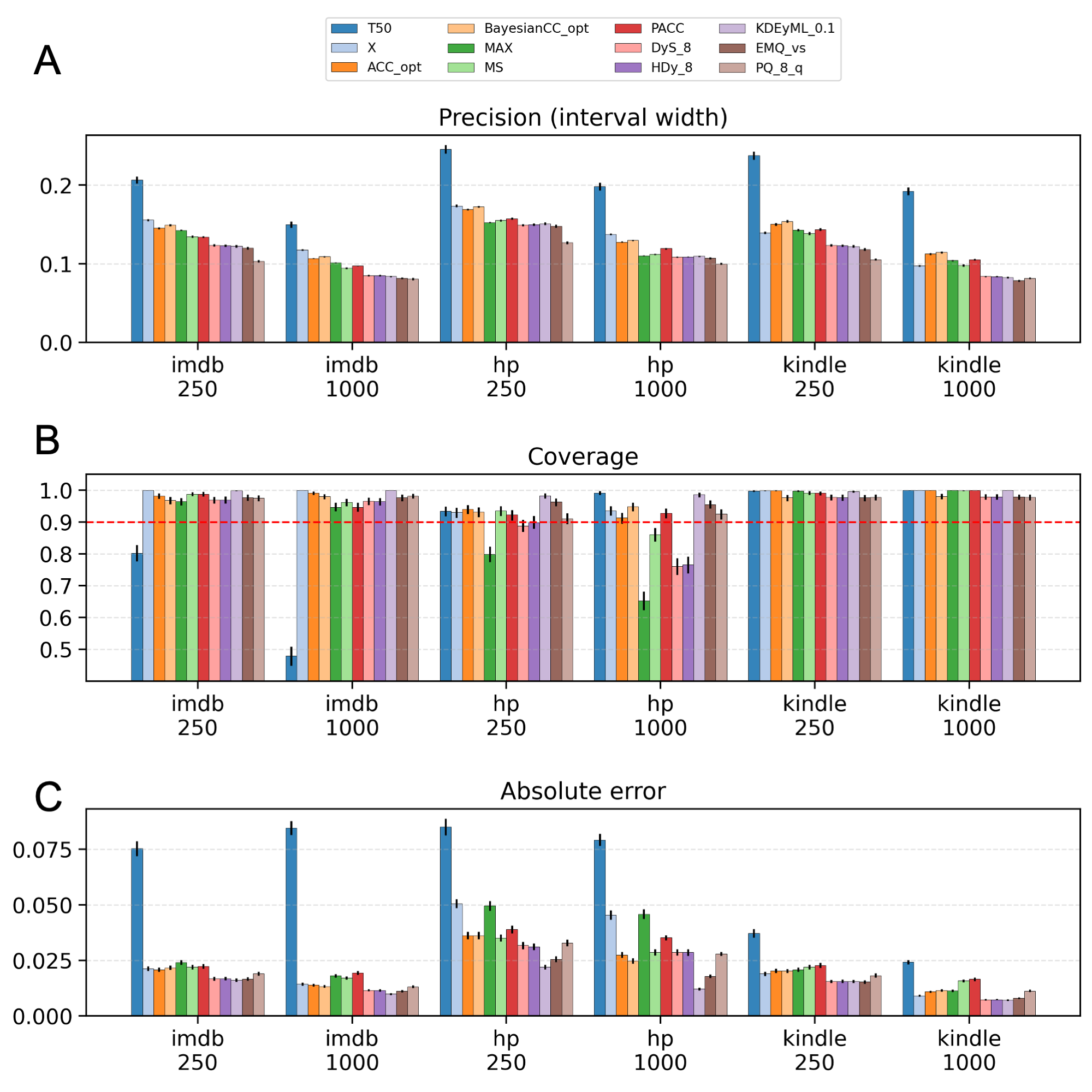}
    \caption{\textbf{Quantification performance on the review datasets}. \small Precision and coverage are for 90\% prediction intervals for PQ, 90\% credible intervals for BayesianCC, and 90\% bootstrap confidence intervals for the other methods. The numbers below imdb, kindle and hp on the x-axis indicate the size of the test set. PQ consistently achieved highest precision without compromising coverage, except for imdb and kindle datasets when validation size and test size were 250 and 1000, respectively. In terms of absolute error, other methods outperformed PQ on the review datasets. Mean values are shown; error bars represent 95\% bootstrap confidence intervals, calculated with 1000 samples. Further details on the datasets and underlying methods are in Sections \ref{experiments} and \ref{appendix}.}
    \label{fig:Fig_S4}
\end{figure}

\newpage
\begin{figure}[!ht]
    \centering
    \includegraphics[width=1\linewidth]{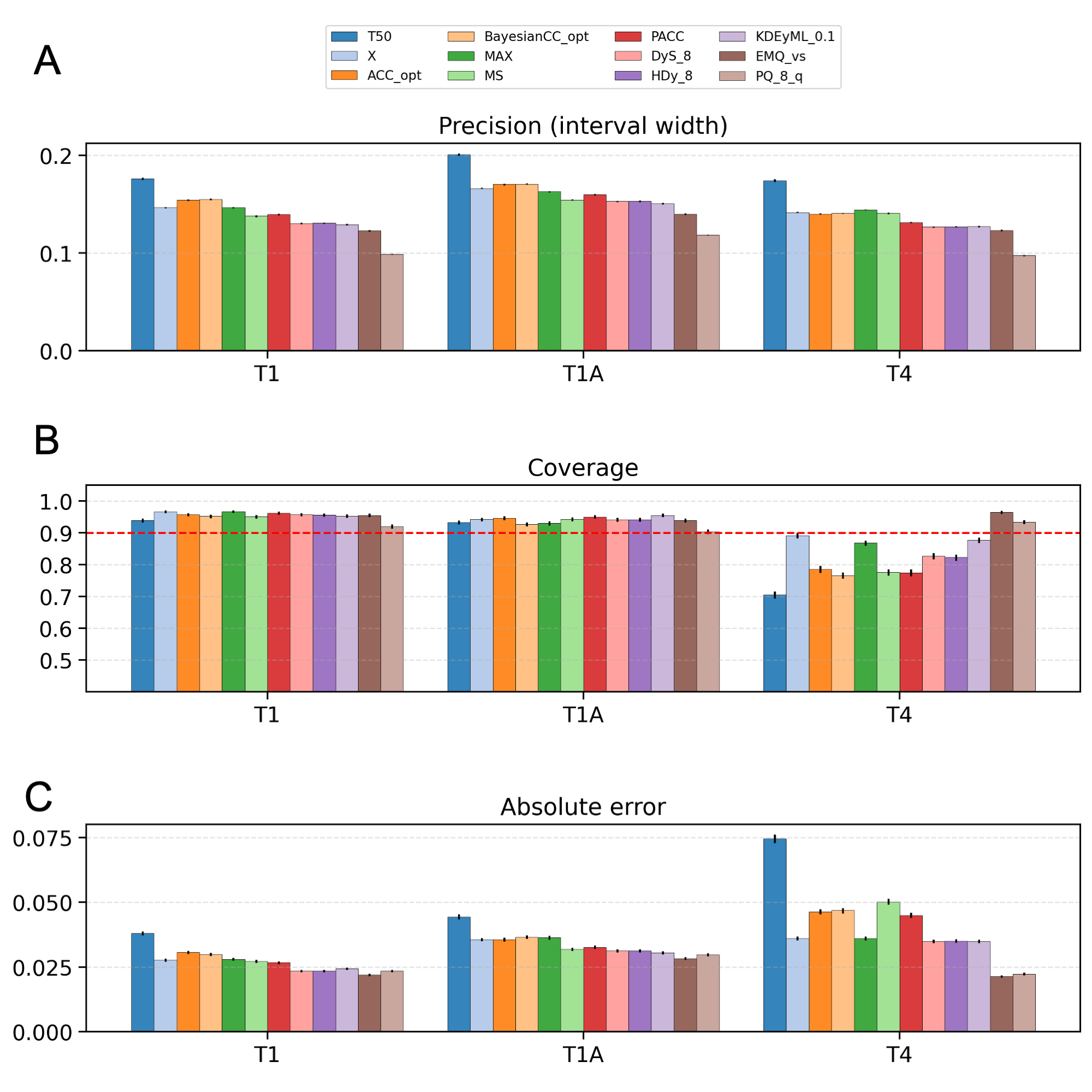}
    \caption{\textbf{Quantification performance on LeQua datasets}. \small Precision and coverage are for 90\% prediction intervals for PQ, 90\% credible intervals for BayesianCC, and 90\% bootstrap confidence intervals for the other methods. The dashed red line in subfigure \textbf{B} indicates the nominal coverage level of 90\%. PQ outperformed other methods in terms of precision on the LeQua datasets. In terms of absolute errors, EMQ outperformed other methods, with PQ being a close second on T4. Mean values are shown; error bars represent 95\% bootstrap confidence intervals, calculated with 1000 samples. Further details on the datasets and underlying methods are in Sections \ref{experiments} and \ref{appendix}.}
    \label{fig:Fig_S5}
\end{figure}

\newpage
\appendix

\section{Technical Appendices and Supplementary Material}
\label{appendix}
\makeatletter\def\@currentlabel{\textit{Appendix}}\makeatother

\label{quantifiers}
In this section, we present the main ideas behind quantifiers against which we compared PQ. We refer to \citet{esuli_learning_2023} for a comprehensive review of existing quantification methods, including those that are non-aggregative. We only considered aggregative methods that are popular in the quantification literature and are already implemented in the QuaPy framework \cite{moreo_quapy_2021}. 

For readability, we maintain the notation already introduced in the previous section. However, the following new notation appears in the description of several quantifiers: we let $f_h(X)$ be a hard classifier ({\it i.e.}, "hard" because the classifier's output is binary, remarking that any soft classifier $f(X)$ can be made hard using some threshold on its output).

\subsubsection{ACC, PACC, MAX, T50, X, and MS}
\label{PACC}
Let $P_T(Y=1)$ be the positive class prevalence in the test set. Let $P_T(f_h(X)=1)$ be the proportion of test data points predicted to be in the positive class. $P_T(f_h(X)=1)$ is the prevalence estimate output by the \textit{classify and count} (CC) method. 

By law of total probability, we have:
\begin{equation}\label{eqn:eqn1}
    P_T(f_h(X)=1) = P_T(f_h(X)=1|Y=1)P_T(Y=1) + P_T(f_h(X)=1|Y=0)(1 - P_T(Y=1))
\end{equation}
Rearranging Equation (\ref{eqn:eqn1}), we get:
\begin{equation}\label{eqn:eqn2}
    P_T(Y=1) = \frac{P_T(f_h(X)=1) - P_T(f_h(X)=1|Y=0)}{P_T(f_h(X)=1|Y=1) - P_T(f_h(X)=1|Y=0)}
    = \frac{P_T(f_h(X)=1) - FPR_T}{TPR_T - FPR_T}
\end{equation}
where $TPR_T$ and $FRP_T$ are respectively the true positive rate and the false positive rate of the classifier on the test set. Nonetheless, these quantities are unknown since true classes for data points in the test set are unknown. Hence, \textit{Adjusted Classify and Count} (ACC) makes the assumption that $P_T(f_h(X)=1|Y=1) = P_V(f_h(X)=1|Y=1)$ and $P_T(f_h(X)=1|Y=0) = P_V(f_h(X)=1|Y=0)$ \cite{forman_quantifying_2008}. That is, TPR and FPR are the same for the validation set and the test set. With this assumption applied to Equation (\ref{eqn:eqn2}), ACC's prevalence estimate becomes:
\begin{equation}\label{eqn:eqn3}
    \hat{P}_T(Y=1) = \frac{P_T(f_h(X)=1) - P_V(f_h(X)=1|Y=0)}{P_V(f_h(X)=1|Y=1) - P_V(f_h(X)=1|Y=0)}
    = \frac{P_T(f_h(X)=1) - FPR_V}{TPR_V - FPR_V}
\end{equation}
where $TPR_V$ and $FRP_V$ are respectively the true positive rate and the false positive rate of the classifier on the validation set.

Because both CC and ACC require binarizing predictions from a soft classifier, we used thresholds that maximized the Matthews Correlation Coefficient (MCC) on the validation data.  We indicate this threshold choice with the "opt" suffix (\textit{i.e.}, ACC\_{opt}). Moreover, we included other threshold selection policies that \citet{forman_quantifying_2006} and \citet{forman_quantifying_2008} proposed, resulting in \textit{MAX}, \textit{T50}, \textit{X}, and \textit{Median Sweep (MS)}. MAX chooses the threshold that maximizes $TPR_V - FPR_V$, the denominator in Equation \ref{eqn:eqn3}; T50 selects a threshold such that $TPR_V \approx 0.5$; X picks a threshold such that $FPR_V = 1 - TPR_V$; MS outputs the median prevalence estimate from an ensemble of methods using all thresholds.

\citet{bella_quantification_2010} proposed the probabilistic version of ACC, called PACC, extending ACC to soft classifiers that output a number in the $[0,1]$ interval, interpreted as a probability of being in the positive class. To maintain the same notation as in Section \ref{theory}, let us define such a classifier as $f(X)$. Therefore, under the same assumptions as ACC, PACC's prevalence estimate is: 

\begin{equation}\label{eqn:eqn4}
    \hat{P}_T(Y=1) = \frac{E_{P_T}[f(X)] - E_{P_V}[f(X)|Y=0]}{E_{P_V}[f(X)|Y=1] - E_{P_V}[f(X)|Y=0]}
\end{equation}

where $E_{P_T}[f(X)]$ is the estimate output by \textit{probabilistic classify and count} (PCC), which, as a counterpart to CC, simply averages predicted class probabilities from the classifier. 

\subsubsection{BayesianCC}

Introduced by \citet{ziegler_bayesian_2024}, BayesianCC is essentially a Bayesian version of the adjusted classify and count (ACC) method discussed in the previous section. We recall that the parameter we aim to estimate is $\theta$, the true positive class prevalence in the test set. We also recall that $n_V$ and $n_T$ are the sizes of the validation and test sets, respectively, while $n^+_V$ and $n^-_V$ respectively represent the sizes of positive and negative subsets of the validation set. Let us also define $\hat{n}^+_T$ and $\hat{n}^-_T$ as the sizes of test data points predicted to be positive by some hard classifier $f_h(X)$. We binarized predictions using thresholds that maximized MCC on the validation data. We indicate this threshold choice with the "opt" suffix (\textit{i.e.}, BayesianCC\_{opt})

Now, let us introduce some new notation related to the confusion matrix of $f_h(X)$. Specifically, by $n^{++}_V$, we will mean the size of the subset of positive validation data points predicted to be positive ({\it i.e.}, true positives). Similarly, by $n^{-+}_V$, we will mean the size of the subset of negative validation data points predicted to be positive ({\it i.e.}, false positives). In other words, the first sign in the superscript refers to the ground truth, while the second sign in the superscript refers to the prediction. Accordingly, we will let $p^{++}=P_V(f_h(X)=1|Y=1)$ and $p^{-+}=P_V(f_h(X)=1|Y=0)$, the true positive rate and false positive rate, respectively. For the binary case we study, BayesianCC then specifies the model as follows:
\begin{gather*}
    n^{++}_V \sim {\rm Binomial}(n^+_V, p^{++})\\
    n^{-+}_V \sim {\rm Binomial}(n^-_V, p^{-+})\\
    \hat{n}^{+}_T \sim {\rm Binomial}(n_T, \theta p^{++}+(1-\theta) p^{-+})\\
    \theta \sim {\rm Uniform(0,1)}\\
    p^{++} \sim {\rm Uniform(0,1)}\\
    p^{-+} \sim {\rm Uniform(0,1)}
\end{gather*}

\subsubsection{HDy, DyS, and KDEyML}
Proposed by \citet{gonzalez-castro_estimating_2010}, HDy's prevalence estimate $\hat{p}$ is found by solving:
\begin{equation}\label{epn:eqn5}
    \hat{p} = \underset{0 \leq p \leq 1}{\arg\min} \ HD(F_T, pF^+_V+ (1-p)F^-_V) 
\end{equation}

where $F_T=P_T(f(X))$, $F^+_V=P_V(f(X)|Y=1)$, $F^-_V=P_V(f(X)|Y=0)$, and $HD$ is the Hellinger Distance between two distributions. In practice, $F_T$, $F^+_V$ and $F^-_V$ are approximated by binning the classifier's output probabilities. We indicate the number of bins used as a suffix (\textit{e.g.}, HDy\_4 and HDy\_8 for 4 and 8 bins, respectively). 

\citet{maletzke_dys_2019} introduced the Distribution \textit{y}\_Similarity (DyS) framework, which generalizes quantifiers in the mixture models family, which HDy is part of. For instance, the DyS framework allows replacing Hellinger distance with Topsøe distance, which improved performance in several cases \cite{maletzke_dys_2019}. In our experiments, by DyS, we mean the use of Topsøe distance in place of Hellinger distance. Similarly to HDy, we indicate the number of bins used as a suffix (\textit{e.g.}, DyS\_8). Finally, similarly to previously described methods, HDy and DyS assume that $P_T(f(X)|Y=1) = P_V(f(X)|Y=1)$ and $P_T(f(X)|Y=0) = P_V(f(X)|Y=0)$.

For both HDy and its DyS generalization, distributions of predicted class probabilities are modeled using histograms. \citet{moreo_kernel_2024} showed that the binning approach is inadequate for the case of multi-class quantification, where inter-class co-variation is important. Thus, \citet{moreo_kernel_2024} proposed using kernel density estimation to model the distributions of class probabilities. This method was named KDEy, and in our experiments we considered its variant called \textit{KDEyML}, which aims to minimize the Kullback-Leibler divergence (instead of Hellinger or Topsøe distance) between validation and test distributions of class probabilities. KDEy methods require setting the bandwidth hyperparameters, and KDEyML was shown to be robust with respect to the hyperparameter \cite{moreo_kernel_2024}. We set the bandwidth to 0.1, the default value in QuaPy. We indicate this choice with a suffix (\textit{i.e.}, KDEyML\_0.1).  

\subsubsection{EMQ}
\label{emq}
The Expectation Maximization Quantifier (EMQ) was proposed by \citet{saerens_adjusting_2002} as a method to recalibrate a classifier's class posterior probabilities when the class prior probability in the test data ({\it i.e.,} the estimate of interest in quantification learning) is different from what it was in the training data. The method is based on the popular Expectation Maximization algorithm for maximum likelihood estimation \cite{dempster_maximum_1977}. The prevalence estimate $\theta$ is calculated via the following iterative process:
\begin{itemize}
    \item At iteration $0$: \hspace{10mm} $\theta^0 = \omega$
    \item At iteration $j$: \hspace{10mm} $f^j(x^T_i)=\frac{\frac{\theta^j}{\omega
    }f(x^T_i)}{\frac{\theta^j}{\omega}{f(x^T_i)} + \frac{1-\theta^j}{1-\omega}(1-f(x^T_i))}$
    \item At iteration $j+1$: \hspace{4mm} $\theta^{j+1} = \frac{1}{n_T}\sum_{i=1}^{n_T} f^j(x^T_i)$
\end{itemize}
where $f^j(x^T_i)$ is the classifier's positive class probability of input $x^T_i$ from the test set after $j$ adjustments; $\omega$ is the estimated source-domain prevalence of the positive class, which is also used as the initialization for the target prevalence estimate. Originally, \citet{saerens_adjusting_2002} proposed that $\omega$ be set to the true class prevalence ({\it i.e.}, positive class prevalence in our case) in the training data. However, \citet{alexandari_maximum_2020} showed that $\omega$ can be set to the average of calibrated predicted class probabilities on a labeled source-domain dataset. We applied the latter approach in our analysis because it does not require access to the data used to train the classifier. That is, we could replace the training set with the validation set by letting $\omega=\frac{1}{n_V}\sum_{i=1}^{n_V} f_{vs}(x^V_i)$, where $f_{vs}(X)$ indicates that class probabilities were calibrated (on the same validation data) using vector scaling (VS) \cite{alexandari_maximum_2020}. We used vector scaling as it was shown to be effective \cite{alexandari_maximum_2020}, which we also confirmed in our exploratory analyses. Our EMQ\_vs notation in this paper indicates the use of vector scaling as the calibration method. By calibrated class probabilities, we mean that those class probabilities reflect the true likelihoods of outcomes \cite{silva_filho_classifier_2023}. A (perfectly) calibrated model outputting 80\% positive class probability for 100 data points implies that 80\% of those data points are actually positive. Poor calibration can deteriorate EMQ's performance \cite{esuli_critical_2021}. 

\end{document}